\documentclass[sigconf]{acmart}
\AtBeginDocument{%
  }

\usepackage{subcaption}
\usepackage{makecell}
\usepackage{multirow} 
\copyrightyear{2026}
\acmYear{2026}
\setcopyright{cc}
\setcctype{by-nc-nd}
\acmConference[COMPASS '26]{ACM SIGCAS/SIGCHI Conference on Computing and Sustainable Societies}{July 27--31, 2026}{Virtual Event, USA}
\acmBooktitle{ACM SIGCAS/SIGCHI Conference on Computing and Sustainable Societies (COMPASS '26), July 27--31, 2026, Virtual Event, USA}
\acmDOI{10.1145/3811242.3819110}
\acmISBN{979-8-4007-2702-3/2026/07} 

\begin{document}

\title{Framing Migration News with LLMs: Structured CoT as a Support for Human Interpretation}

\author{David Alonso del Barrio}
\authornote{Both authors contributed equally to this research.}
\email{ddbarrio@idiap.ch}
\affiliation{%
  \institution{Idiap Research Institute}
  \country{Switzerland}
}

\author{Jing Wen}
\authornotemark[1]
\email{jing.wen@etu.unige.ch}
\affiliation{%
  \institution{Idiap Research Institute and University of Geneva}
  \country{Switzerland}
}

\author{Daniel Gatica-Perez}
\email{gatica@idiap.ch}
\affiliation{%
  \institution{Idiap Research Institute and EPFL}
  \country{Switzerland}
}









\renewcommand{\shortauthors}{Alonso del Barrio et al.}

\begin{abstract}
\footnote{\textbf{David Alonso del Barrio, Jing Wen, Daniel Gatica-Perez| ACM 2026. This is the author's version of the work. It is posted here for your personal use. Not for redistribution. The definitive Version of Record will be published in ACM SIGCAS/SIGCHI Conference on Computing and Sustainable
Societies (COMPASS'26) https://doi.org/10.1145/3811242.3819110}}
Frame analysis of migration news is a socially consequential task: media scholars and researchers who study how migration is narrated need tools that are not only accurate, but transparent, auditable, and accessible within the resource constraints typical of academic research groups. Existing LLM-based approaches rely on proprietary APIs and large models that raise concerns about data privacy, reproducibility and equitable access among media researchers. This work studies how a locally 
deployable open-source LLM can support interpretable frame analysis as an assistive tool. We introduce a Structured Chain-of-Thought (SCoT) prompting approach using Llama3-8B, enabling step-by-step justifications grounded in predefined framing categories. This structured design allows users to audit model outputs and examine alternative interpretations in a task that is inherently subjective. We evaluate our approach on a dataset of migration-related news and show that SCoT improves classification performance over zero-shot and few-shot baselines while remaining feasible on a single GPU. Then, we conduct a human-centered evaluation in which annotators assess the coherence and influence of ``the model's reasoning''. Results indicate that SCoT explanations are generally perceived as logical (mean score 4.1/5, though with notable variation across texts) and can prompt reflection on initial interpretations, even when disagreement persists. Our findings highlight both the potential and risks of LLM-assisted frame analysis. While structured reasoning can increase the traceability of model outputs and support critical interpretation, it can also influence human judgment in subtle ways. By enabling local deployment and emphasizing human-in-the-loop interaction, this work contributes to discussions on responsible and accessible computational tools for the study of socially impactful media narratives.
\end{abstract}



\maketitle

\section{Introduction}

A core function of frames is to guide public interpretation of events by emphasizing certain elements. For example, a large-scale protest could be framed by emphasizing the potential economic disruption it causes through business closures and lost productivity (economic frame). Alternatively, media coverage might highlight the security concerns and the need for increased police presence to maintain order (security and defense frame). Finally, the protest could be portrayed by focusing on the political motivations of the organizers and its potential impact on upcoming elections or policy debates (political frame). These frames do not merely describe events, they influence emotional reactions, political attitudes, and policy preferences~\cite{druckman2001evaluating, chong2007framing, nelson1997media}.
In domains such as migration, these framing processes are particularly important, as media narratives can shape public discourse and inform policy debates around socially sensitive and contested issues \cite{seiger2025navigating}. Understanding how such interpretations are constructed and analyzed is therefore central to research in communication, sociology, and related fields.

With the emergence of LLMs, automated frame classification via prompt-based techniques is being studied ~\cite{ alonso2023framing,  weinzierl2024discovering, alizadeh2025open}. While results show that models can reach agreement rates comparable to crowd workers, especially when fine-tuned, most evaluations focus narrowly on agreement rate. 
Less attention is paid to interpretability, specifically whether the model's ``reasoning'' is understandable to humans, how it is influenced by data, and how it might shape human judgment in subjective tasks like frame analysis. 

Existing models often require powerful hardware or external APIs, unviable in resource-constrained environments common in critical media analysis (e.g., academic research groups, independent media organizations). These settings not only face infrastructure limitations—such as running on a single GPU—but also raise ethical concerns, such as unauthorized use of journalistic content and potential privacy leakage when training commercial models~\cite{le2022trust, tseng2025ownership}. Such constraints can limit access to computational tools for researchers studying media in contexts where understanding framing is socially relevant.

To address these challenges, this study focuses on two research questions:

\textbf{RQ1:} To what extent can structured prompting strategies provide interpretable frame classification with resource-constrained LLMs, and how does SCoT facilitate this process?

\textbf{RQ2:} How do human evaluators perceive and respond to the model ``reasoning'' in SCoT-based frame classification, particularly in terms of agreement and confidence, given the task's inherent subjectivity?

Our research addresses these questions with the following contributions:

\textbf{1}. We introduce a structured Chain-of-Thought (SCoT) prompting approach for interpretable frame identification in news text using a local Large Language Model (Llama3-8B). This deductive approach, where the model selects from a predefined set of 14 frames, enhances transparency by providing step-by-step ``reasoning'', including justifications grounded in frame definitions. Implemented on a single GPU to simulate resource-constrained environments, SCoT addresses critical challenges in the media industry such as the costs of external AI infrastructure, reliance on opaque black-box models, and data privacy concerns associated with external API usage \cite{simon2024artificial}. By enabling local and interpretable analysis, this approach could support more accessible use of computational methods in media research.
    
\textbf{2}. We conducted a human evaluation task where 10 annotators, after independently selecting frames and highlighting supporting evidence in 10 news texts about migration, assessed the coherence and potential validity of the ``reasoning'' provided by our SCoT-driven model. 
Our findings indicate that even when the model's final frame classification differed from the initial human annotation, the generated arguments were often considered sensible, and highlighted plausible alternative interpretations not initially considered by the annotators (achieving a score of 4.1 out of 5 in the evaluation of LLM's ``reasoning''). This points toward a potential assistive role for resource-efficient LLMs in subjective classification tasks like frame analysis, potentially enriching human understanding and offering diverse perspectives.

Together, these contributions speak to broader questions around the responsible design of computational tools for socially sensitive research. Migration is not only a contested media topic, since it is explicitly recognized in the UN's 2030 Agenda: SDG 10.7 calls for facilitating orderly and safe migration, while SDG 16.10 mandates public access to information and protection of fundamental freedoms \cite{un2015agenda}. Prior work has shown that media framing of migration is relevant to monitoring these goals, as the narratives that dominate coverage can either support or undermine public understanding of migration as a shared societal challenge \cite{fernandez2024migration}. By prioritizing local deployment, interpretability and human oversight over raw classification performance, this work aligns with ongoing discussions on how AI tools can be made more accessible and accountable to the research communities engaged in this kind of analysis \cite{ollion2024dangers, tornberg2024best}.



The rest of this work is structured as follows. Section \ref{sec: related_work} reviews prior work on news frame analysis and computational methods. Section \ref{sec: data} describes the dataset used. Section \ref{sec: methodology} outlines the methodology. Section \ref{sec: results and discussion} presents the results and their discussion. In Section \ref{sec: FATE}, we discuss the sociotechnical implications of the use of LLMs in frame analysis, and then we present the limitations of our research in Section \ref{sec: Limitations}. Finally, Section \ref{sec: conclusion}  concludes the study by answering the research questions.

\section{Related Work} 
\label{sec: related_work}
In media studies, \citeauthor{goffman1974frame} originally defined frames as cognitive patterns for interpreting social events \shortcite{goffman1974frame}. Building on this micro-sociological foundation, \citeauthor{entman1993framing} expanded the concept to political discourse and mass communication, defining framing as \textit{``the selection of certain aspects of perceived reality and making them more salient in a communicative text to facilitate a particular problem definition, causal explanation, moral evaluation, and/or treatment recommendation.''}~\shortcite{entman1993framing}.
Within frame analysis, \citeauthor{sullivan2023three} identified three levels; semantic framing (language-based), cognitive framing (thought-based) and communicative framing (communication-based) \shortcite{sullivan2023three}. To understand these three aspects, we are going to present an example around the topic of migration. The semantic level focuses on the specific words used to describe an issue, for example, referring to migrants as ``refugees'' versus ``illegal aliens'', where the first one evokes empathy, suggesting that there are people in danger, while the second one emphasizes illegality, evoking fear or distrust. The cognitive aspect focuses on how the audience interprets the terms based on previous knowledge. For example, a person who values humanitarian aid might associate ``refugees'' with vulnerability and the need for help, while a person concerned about national security might interpret ``illegal aliens'' as a threat to national borders. Finally, the communication aspect concerns how the frame is repeatedly communicated to influence public opinion. There may be media or politicians who repeatedly refer to migrants as invaders in headlines or speeches, while at the same time there may be campaigns focusing on asylum seekers and emphasizing stories of people leaving countries in war. These three levels are interconnected and, depending on the semantics in the communication, trigger one cognition or another. 
Communication, therefore, contains semantics plus cognition.

In addition to these three levels, media framing research has generally developed along two major methodological trajectories: deductive \cite{dirikx2010frame} and inductive \cite{van2010strategies}. The deductive approach is based on the identification in the text of one or more predefined frames, while the inductive approach allows the frames to emerge organically.
At the same time, scholars have also distinguished between generic frames \cite{semetko2000framing,boydstun2014tracking}, which apply across domains, and issue-specific frames, which are tailored to a particular topic \cite{wicke2020framing, liu2019detecting}.

From an NLP perspective, frame identification has been widely studied \cite{ali2022survey,otmakhova2024media}.
\citeauthor{otmakhova2024media}\cite{otmakhova2024media} identified four types of media framing relevant to one or more of the levels described above and examined their coverage in NLP approaches. These types are: ``emphasis framing'', which consists of highlighting certain aspects and excluding others when presenting information, aligning with the definition of Entman \shortcite{entman1993framing}. Then, there is ``equivalency framing'', where the information is classified in a positive or negative way \cite{druckman2004political}; ``framing by word choice and labeling'', where the frame is more focused on the semantic level, and the selection of certain vocabulary to present the information \cite{pearson2010undocumented}; and finally ``narrative framing'', which focuses on abstract devices, syntactical and rhetorical structures when presenting a piece of information \cite{jones2010narrative, frermann2023conflicts}. We focus on emphasis framing, building on the previous work of frame analysis with LLMs that we discuss next.

Survey papers \cite{ali2022survey,otmakhova2024media} raise concerns about the automatization of frame analysis, arguing that many approaches tend to identify (sub)topics rather than actual frames. This is reflected in the widespread use of topic modeling as a predominant technique for frame identification \cite{dimaggio2013exploiting, sarmiento2022identifying}.

The recent advent of LLMs presents a significant shift, offering novel avenues for frame classification in NLP. We can identify three main prompt-engineering techniques for frame classification: zero-shot (task without examples \cite{xian2018zero}), few-shot (task with few examples \cite{brown2020language}), and Chain-of-Thought (CoT, ``reasoning'' through intermediate steps \cite{wei2022chain}).

Alonso del Barrio et al. \shortcite{alonso2023framing, alonso2024human} used GPT-3.5 in a zero-shot setting to classify headlines and TV transcripts using \citeauthor{semetko2000framing} 
\shortcite{semetko2000framing} five generic frames, showing promising results for using LLMs in frame identification, because despite the subjectivity of the task, the model chose frames similarly to human annotators.   \citeauthor{gilardi2023chatgpt} \shortcite{gilardi2023chatgpt} classified the frames of tweets based on the 15 generic frames of the Media Frame Corpus (MFC) using ChatGPT and a zero-shot prompting approach. Then,  \citeauthor{alizadeh2025open} \shortcite{alizadeh2025open} used the same approach but with open-source LLMs and adding few-shot prompting and Chain-of-Thought, showing that the LLMs perform better than crowd-workers in annotation tasks.
While prior work has evaluated the performance of LLMs (open source and not) 
on frame classification tasks, most of the evaluations remain centered on model agreement rate. Fewer studies have examined how the model’s internal ``reasoning'' is perceived or understood by human users. Moreover, although some recent works have incorporated Chain-of-Thought (CoT) reasoning into text (tweet) classification pipelines~\cite{alizadeh2025open}, these efforts primarily report performance metrics, without deeply analyzing how persuasive or plausible the ``reasoning'' appears to human users. Complementing these methodological evaluations, recent interactive platforms such as the Media Bias Detector \cite{wang2025media} have begun to operationalize large-scale analysis of news framing and political lean using proprietary LLMs. While such tools demonstrate the potential for real-time bias detection, their reliance on proprietary, ``black-box'' APIs has highlighted a critical need for greater transparency and local reproducibility in academic media research. 

In the work of \citeauthor{weinzierl2024discovering} \shortcite{weinzierl2024discovering}, they used Chain of Thought (CoT) to identify frames in social media posts in an inductive way. The authors affirm that previous literature is focused on the identification of the problem, 
but not on the causal interpretation, which they called articulation (expressing the reason or causes of salient problems). We partially disagree with this view. 
While predefined frames, such as the 15 generic frames proposed by \citeauthor{boydstun2014codebook} \shortcite{boydstun2014codebook} or the 5 frames by \citeauthor{semetko2000framing} \shortcite{semetko2000framing}, may describe problems or topics, they can also function as explanations for the underlying causes of these issues. For example, we consider the Fairness and Equality Frame: in one instance, a headline like ``Migrant Workers Demand Equal Pay and Working Conditions'' demonstrates the frame as a subtopic/problem. Conversely, ``Unequal Treatment of Migrants in the Labor Market Fuels Tensions Over Workplace Equality'' exemplifies the frame as a causal interpretation of workplace tensions, so the predefined frames can function as both problem descriptors and causal interpreters. Despite this disagreement, we agree on the need for explicit evidence in texts to support frame selection, and we posit that CoT, with its capacity to generate ``reasoning'' steps, offers a promising approach. In contrast to \citeauthor{weinzierl2024discovering} \shortcite{weinzierl2024discovering} inductive methodology applied to social media posts, we employed a deductive strategy using predefined frames from \citeauthor{boydstun2014codebook} \shortcite{boydstun2014codebook} and analyzed longer excerpts (around 250 words) from articles. The fact that our approach is deductive with a predefined list of frames allows us to compare the results with both other topics and the previous literature. Additionally, our approach included human evaluation to assess the quality of the generated arguments, focusing on how well the ``reasoning'' supports interpretation in a subjective task and how it is perceived by individuals engaging with the analysis.

\section{Data} 
\label{sec: data}
This research uses a case study of immigration news, a topic highly relevant for frame analysis due to its potential social implications \footnote{While our study focuses on migration in general, we will use the term ``immigration'' in this section and subsequent discussions to align with the nomenclature of the original dataset.}. We used the articles about immigration of the MFC~\cite{card2015media}, where each sample contains four components: a news text (~250 words on average), a primary frame annotation selected from 15 predefined frame categories derived from the Policy Frames Codebook~\cite{boydstun2014codebook}, the newspaper, and the date of the text. The annotation of the primary frame was conducted by undergraduate students at a U.S. research university. The annotation process consisted of three phases. In the first phase, the annotators independently annotated a series of items from the definitions. In a second stage, in pairs, they annotated and discussed with the partners those cases where they tended to have more disagreements, and finally in a third stage, the pairs annotated and in case of disagreement they discussed until an agreement was reached. The authors emphasized that there was subjectivity in the task and even if an agreement was reached, it did not mean that the discarded option was incorrect. 
In Table~\ref{tab:frame_definitions}, we present the definition for each frame. 

\begin{table}[h!]
\setlength{\tabcolsep}{1mm} 
\small 
\centering
\begin{tabular}{p{3.3cm}p{5cm}} 
    \hline
    \textbf{Frame Category} & \textbf{Definition} \\
    \hline
    Economic & ``Costs, benefits, or other financial implications.'' \\\hline
    Capacity and resources & ``Availability of physical, human, or financial resources, and capacity of current systems.'' \\\hline
    Morality & ``Religious or ethical implications.'' \\\hline
    Fairness and equality & ``Balance or distribution of rights, responsibilities, and resources.'' \\\hline
    \makecell[l]{Legality, constitutionality,\\and jurisprudence} & ``Rights, freedoms, and authority of individuals, corporations, and governments.'' \\\hline
    Policy prescription and evaluation & ``Discussion of specific policies aimed at addressing problems.'' \\\hline
    Crime and punishment & ``Effectiveness and implications of laws and their enforcement.'' \\\hline
    Security and defense & ``Threats to the welfare of the individual, community, or nation.'' \\\hline
    Health and safety & ``Health care, sanitation, public safety.'' \\\hline
    Quality of life & ``Threats and opportunities for the individual’s wealth, happiness, and well-being.'' \\\hline
    Cultural identity & ``Traditions, customs, or values of a social group in relation to a policy issue.'' \\\hline
    Public opinion & ``Attitudes and opinions of the general public, including polling and demographics.'' \\\hline
    Political & ``Considerations related to politics and politicians, including lobbying, elections, and attempts to sway voters.'' \\\hline
    External regulation and reputation & ``International reputation or foreign policy of the U.S.'' \\\hline
    Other & ``Any coherent group of frames not covered by the above categories.'' \\
    \hline
\end{tabular}
\caption{Framing dimensions, originally proposed in Boydstun et al. \cite{boydstun2014codebook} and then used in Card et al. \cite{card2015media}}
\label{tab:frame_definitions}
\end{table}

\begin{figure}[htbp]
    \centering
    \includegraphics[width=\linewidth]{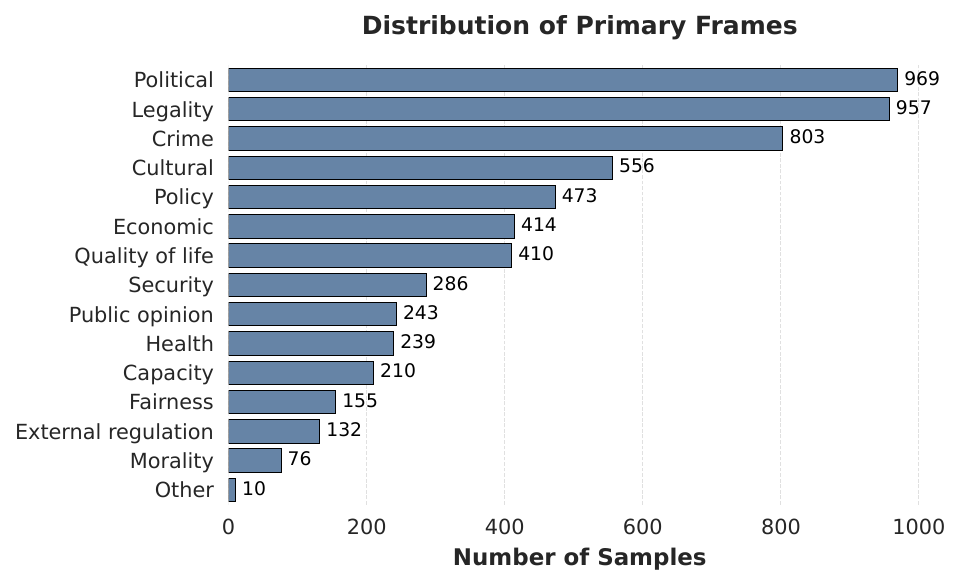}
    \caption{Distribution of frames in the MFC immigration subset. Label names in this figure have been shortened for clarity; refer to Table \ref{tab:frame_definitions} for complete names.}
    \Description{The figure, titled ``Distribution of Primary Frames'', is a horizontal bar chart displaying the frequency of various frames identified within the MFC immigration subset. The y-axis lists the frame categories in descending order of frequency, ranging from ``Political'' (969 samples) at the top to ``Other'' (10 samples) at the bottom. The x-axis represents the ``Number of Samples'' on a scale from 0 to 1000. Each category is represented by a blue bar with the corresponding numerical count displayed at its end. The chart caption notes that the label names have been shortened for clarity and refers readers to Table 1 for complete definitions}
    \label{fig:immigration_distribution}
\end{figure}

There are 5,933 articles related to immigration with a clear imbalanced distribution of frames, visualized in Figure~\ref{fig:immigration_distribution}. 

We excluded the rarely used ``Other'' frame, which lacks a specific representation, to improve result interpretability. 
For analysis, we randomly sampled a representative subset of 700 news articles (11\% of the total dataset), retaining its original imbalanced distribution for manageable computational load and reflecting key characteristics. Additionally, we created a balanced version by selecting 50 texts from each of the remaining 14 frame categories, totaling 700 samples. We consider 700 texts a sufficiently large sample to evaluate our SCoT proposal. Furthermore, having both balanced and imbalanced versions allows us to understand the influence of data distribution on agreement rate results.

\section{Methodology}
\label{sec: methodology}
This section outlines our two-part methodology: (1) evaluating a single-GPU deployable LLM for interpretable frame identification, and (2) a human evaluation task assessing SCoT-generated ``reasoning'', to understand consistency with human interpretations.

\subsection{Model Selection}
The selection of the model was based on a balance between performance, computational feasibility, and task suitability. Llama models have gained significant traction in NLP tasks, particularly in classification and reasoning tasks, due to their open-access nature and efficiency compared to proprietary alternatives \cite{touvron2023llama}.  Llama3, 
the latest version when our research was conducted, introduced key enhancements over Llama2, including an expanded vocabulary, a larger training dataset, and improved instruction-following capabilities \cite{meta2024llama3}. From available sizes (1B, 3B, 8B, 11B, 70B, 90B, 405B), we excluded: 11B/90B (multi-modal, not text-focused); 1B/3B (struggled with long texts and complex frame definitions in preliminary tests, yielding nonsensical/incomplete outputs); and 70B/405B (impractical for single-GPU due to computational demands). 
As a result, we selected Llama3-8B, balancing computational feasibility and expected performance. 

Regarding implementation details, the local experiments were conducted on a single GPU RTX-3090. We used the Transformers pipeline of Hugging Face \cite{wolf2020transformers} to implement and run Llama3-8B on a local environment. 

\subsection{SCoT Design}
To enhance interpretability and understand LLM logic in frame identification, we employed Chain-of-Thought (CoT) prompting \cite{wei2022chain}. CoT enables complex ``reasoning'' through intermediate steps, making model predictions more transparent, and allowing comparison with human interpretations.

Initial CoT designs, like the ``let’s think step by step'' approach \cite{kojima2022large}, lacked uniformity, hindering systematic evaluation and producing variable output structures. Our exploratory analysis identified inconsistent patterns (e.g., summary, frame discussion, selection) and verbosity. Inspired by \citeauthor{li2025structured}'s SCoT for code generation \cite{li2025structured}, we adapted a Structured CoT (SCoT) for frame analysis. This approach guides the model through a predefined sequence of steps, ensuring uniformity and transparency in ``reasoning''. SCoT improves interpretability, aids error analysis, and offers more focused, efficient output by limiting sentences per step.

\begin{figure}[h!]
    \centering
    \includegraphics[width=\linewidth]{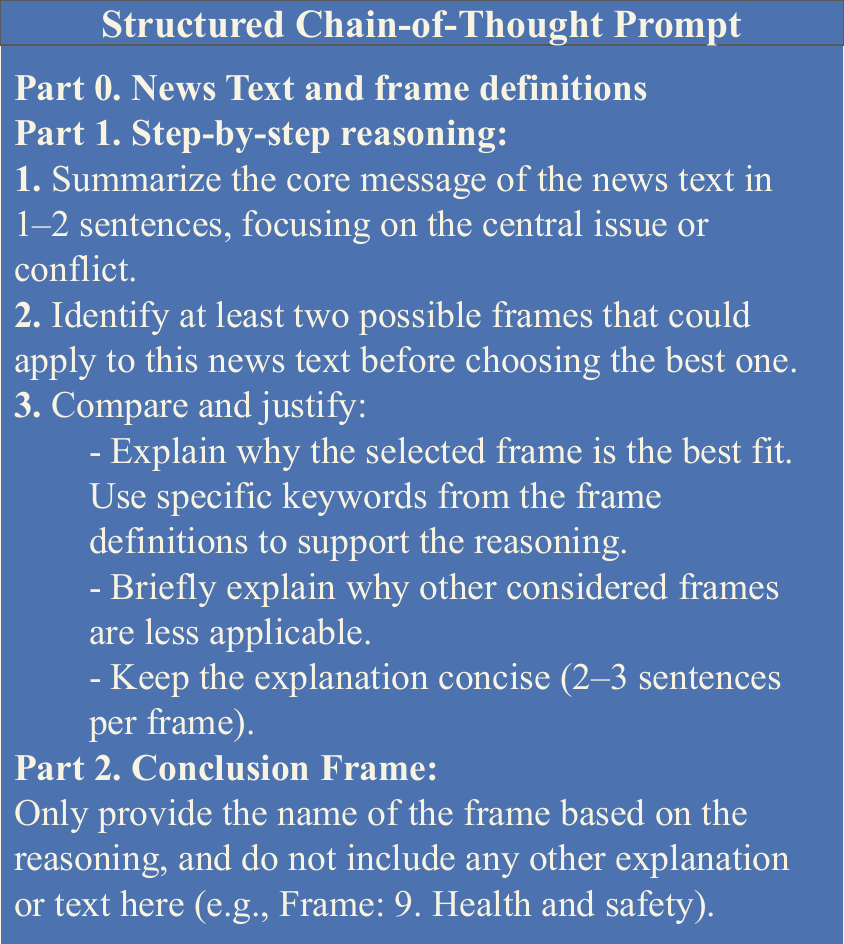}
    \caption{SCoT prompt.}
    \Description{The figure titled ``SCoT Prompt'' presents a hierarchical, text-based template for analyzing news texts, organized against a solid blue background. The prompt begins with ``Part 0'', designated for inputting news text and frame definitions, followed by ``Part 1'', which outlines a three-step reasoning process: summarizing the core message in one to two sentences, identifying at least two potential frames, and providing a justification that explains the best-fit frame using specific keywords while briefly contrasting it with other considered frames. The template concludes with ``Part 2'', which instructs the LLM to provide only the name of the final chosen frame without including any additional explanation or supplementary text.}
    \label{fig:Structured_CoT}
\end{figure}

 As shown in Figure~\ref{fig:Structured_CoT}, the prompt is divided into three parts: (0) news text and frame definitions, (1) ``reasoning'' steps that progressively analyze and compare possible frames, and (2) an explicit classification conclusion output.

To evaluate performance, we apply the following classification metrics:
\begin{itemize}
    \item Agreement Rate: This measures whether the final predicted frame matches the dataset label.
    \item F1-score: This is the harmonic mean of precision and recall across all 14 categories.
    \item Confusion matrix: This is a visual representation of per-class prediction errors, highlighting which frame categories are most frequently confused with one another. This allows us to interpret common ``reasoning'' errors and assess semantic overlaps between frames.
\end{itemize}

\subsection{Human evaluation of LLM ``Reasoning''}
To complement classification metrics, we conducted a human evaluation task to assess the coherence and interpretability of SCoT-generated ``reasoning'' from Llama3-8B. The goal of the human evaluation is not to establish a new ground truth for news frames or to measure inter-annotator agreement. Instead, we evaluate whether SCoT-based explanations make LLM-assisted frame classification more transparent and interpretable to human users. Annotators are asked to engage with model outputs as users of an annotation assistant, assessing the coherence and usefulness of the generated reasoning rather than converging on a single correct label.
\subsubsection{Annotators Recruitment and Setup}
We recruited 10 annotators via an open call, selected for their familiarity with news text analysis. The majority were graduate students in digital humanities with backgrounds in media studies and AI, as well as graduate students with a background in NLP. For the evaluation we created an interactive environment using LimeSurvey \cite{limesurvey} to do the annotation, and it took about 90 minutes. Annotators were compensated for their time (the equivalent of \$25 USD per hour, for two hours). The structure of the annotation process was as follows:
\begin{itemize}
    \item Introduction: In this section, we first presented the concept of frame analysis in theoretical terms, and then, using various examples of headlines and news stories that we showed on the screen, the annotators decided and discussed what the main frame associated with that headline or news story was.
    \item Practice Task: One warm-up question (Text 0) to familiarize annotators with the annotation interface and task requirements. During this phase, annotators could ask clarification questions.
    \item Main Task: Independent annotation of 10 immigration-related news texts, with no external assistance allowed. The selection of the 10 texts was not random. We decided to choose 10 texts that had different labeled frames. Of those 10 cases, in 5 of them the LLM argument supported the dataset label, while in the other 5 cases it gave a different alternative. In this way, we ensured variety in the evaluation of possible cases, rather than simply choosing cases where there was already agreement between the LLM and the original dataset annotation where the examples were very obvious. This information was never shared with the annotators so that it would not influence their evaluation.
\end{itemize}
We acknowledge that 10 annotators and 10 texts constitute a small-scale, exploratory evaluation. In line with formative HCI study design \cite{schroeder2025large}, our aim is not statistical generalizability but to surface qualitative patterns in how structured LLM reasoning interacts with human interpretation. The study is intentionally scoped to generate hypotheses and design insights for future work, rather than to establish definitive quantitative conclusions.
\subsubsection{Question Design and Response Flow}
The survey was divided into two sequential sections to prevent annotators from adjusting their initial frame choice after reading the LLM-generated ``reasoning''. We refer to the survey questions as follows:
``QX'' means ``Question X''.

\paragraph{(a) Phase 1: Initial Frame Selection and Justification (Q1–Q2).}
    
    \begin{itemize}
        \item Frame classification (Q1):
        ``Read text X of the news article below and choose a news frame from the 14 options.'' 
        
        The goal was to establish a baseline of independent human interpretation, assessing inherent subjectivity, and comparing agreement levels across original human annotations, our human annotations, and the LLM. 
        \item Justification (Q2):
        ``What sentences from the text support your choice of frame? (You can copy phrases from the original text or write your own explanation.)'' 
        
        This aimed to ensure choices were text-supported.
    \end{itemize}
    
    When annotators answered Q1 and Q2, they were locked out from modifying their initial answers, ensuring that any influence from LLM-generated ``reasoning'' would only be captured in the next phase.
    
\paragraph{(b) Phase 2: Exposure to LLM ``Reasoning'' and Reconsideration (Q3–Q5).}
    
    After submitting their initial frame selection, annotators were presented with the ``reasoning'' of the model and asked to evaluate it.
    \begin{itemize}
        \item Evaluation of LLM ``reasoning'' (Q3): 
        ``Here is a structured reasoning approach used to justify the selected frame. Reasoning Analysis for Text X: ... Text X (in case you want to consult it again): ... Do you think the arguments used to choose the given frame make sense? Scale 1-5 (Definitely no - Probably no - Maybe - Probably yes - Definitely yes)'' 
        
        The goal was to evaluate whether human annotators perceive the SCoT ``reasoning'' as logical and coherent. If the ``reasoning'' of the LLM received high scores, even if the LLM’s frame choice did not match the human’s, it suggested that structured ``reasoning'' enhanced interpretability and made sense. Conversely, low ratings might indicate inconsistencies in model ``reasoning'' or misalignment with human cognition.
        
        \item Frame Reconsideration (Q4):
        ``If the primary frame you selected was different from the frame shown above, would you change your choice after reading the provided arguments? (Scale 1-5: 1 = Definitely no, 5 = Definitely yes). If you chose the same frame, select the default option `No answer' ''. 
        
        This assessed the potential persuasiveness of SCoT ``reasoning''; frequent revisions might suggest a certain impact on human decision-making.
        \item Justification of change (Q5):
        ``If you would change your framing choice, explain why. If you would not change your framing choice, explain why. If you are unsure, explain why. You may reference phrases from the original text or the structured ``reasoning'' analysis. If you chose the same frame, simply write `the same frame'.'' 
        
        With this question, we aimed to understand why humans either accept or reject LLM-generated ``reasoning''. By analyzing the responses, we investigated whether changes in classification were due to persuasive model ``reasoning'', or the recognition of multiple valid frames in complex texts. This concern is salient given that LLMs may produce fluent yet misleading outputs,  which appear convincing despite lacking grounding in human reasoning~\cite{bender2021dangers}.
        \end{itemize}

 With this small-scale evaluation, we focused on how annotators interpret and respond to the SCoT-generated explanations, rather than treating interpretability as a purely model-side feature. Even if the agreement rate between humans and the model is low, the ``reasoning'' process may still hold value for assisting in frame classification tasks, particularly in complex multi-frame texts. 

\section{Results and Discussion}
\label{sec: results and discussion}
In this section, we present and discuss the results of our applied methodology.

\subsection{SCoT Prompting: Performance and Interpretability (RQ1)}
\label{subsec:structured CoT}
This subsection evaluates Llama3-8B's performance locally, specifically analyzing SCoT's impact on frame classification and interpretability.


\subsubsection{Dataset Distribution and Classification Performance}

Here, we examine how label distribution influences prompt performance, highlighting real-world media challenges. Table~\ref{tab:cot-results} puts SCoT prompt results in context on balanced and imbalanced datasets. In addition to SCoT, we performed the frame analysis exercise with zero-shot learning and a representative few-shot baseline (5 examples) to contextualize the improvement introduced by SCoT. We decided to discard an ``unstructured'' CoT design, like the ``let’s think step by step'' approach \cite{kojima2022large} because it often failed to reach a conclusion and was too verbose, whereas SCoT had the same structure throughout and allowed for systematic evaluation. 
\begin{table}[!htp]
\centering
\resizebox{\columnwidth}{!}{ 
\begin{tabular}{ccccc}\toprule
Metric &Method &Balanced Dataset &Imbalanced Dataset \\\midrule
\multirow{3}{*}{Agreement} &Zero shot learning &28\% &40\% \\
&Few shot learning 5 examples &28\% &41\% \\
&SCoT &\textbf{36\%} &42\% \\\hline
\multirow{3}{*}{F1-score} &Zero shot learning &0.19 &0.20 \\
&Few shot learning 5 examples &0.24 &0.27 \\
&SCoT &\textbf{0.28} &0.29 \\
\bottomrule
\end{tabular}
}
\caption{Performance comparison across balanced and imbalanced datasets. Balanced results are the primary benchmark for a rigorous 14-class evaluation. Note that absolute figures are modest by design: Llama3-8B is used here not as a state-of-the-art classifier but as a locally deployable, interpretable assistant. The key contribution is the SCoT output structure, which enables the human-centered evaluation in Section 5.2.}

\label{tab:cot-results}
\end{table}
The imbalanced dataset shows a higher agreement rate for all the prompt engineering techniques, suggesting that higher agreement largely reflects the dominance of frequent categories rather than improved discrimination across frames. The F1-score, a more reliable classification metric, remains stable between balanced and imbalanced settings, indicating no significant performance improvement despite the higher agreement rate. We emphasize results on the balanced dataset, which provide a more rigorous assessment of performance across all 14 frames. Notably, our SCoT approach reaches 36\% agreement and 0.28 macro F1 on the balanced dataset, outperforming the 5-shot baseline (28\% agreement, 0.24 F1) and zero-shot (28\% agreement, 0.19 F1). While these absolute figures are modest (as expected for a 14-class subjective task on a single 8B model without fine-tuning), the primary value of SCoT in this work is not classification performance per se, but the structured reasoning it produces. It is the traceability and auditability of SCoT outputs, rather than their accuracy, that enables the human-centered evaluation we conduct in Section 5.2, and that makes this approach relevant for researchers who need to understand, critique and validate model ``reasoning'' in sensitive social domains.

As a comparison, previous work has primarily reported agreement rates, often without explicitly analyzing the influence of dataset distribution. For instance, \citeauthor{alizadeh2023open}  \shortcite{alizadeh2023open, alizadeh2025open} report average agreement rates for ChatGPT and open-source models (40-50\%), but do not distinguish performance by class distribution. The same applies to the work of Alonso del Barrio et al. \shortcite{alonso2023framing, alonso2024human} (43-49\% of agreement rate) where they compare their results with previous literature in terms of agreement rate, without 
further analyzing the reasons behind such rates. Our results highlight the importance of interpreting agreement metrics in the context of dataset composition. Especially for tasks influenced by subjectivity like frame classification, agreement alone may not reflect true model performance, and should be complemented with metrics such as F1-score, suitable for cases of imbalanced data. 

\subsubsection{Frame-Level Confusion and Interpretive Ambiguity}
\begin{figure}[h]
  \centering
  \includegraphics[width=\linewidth]{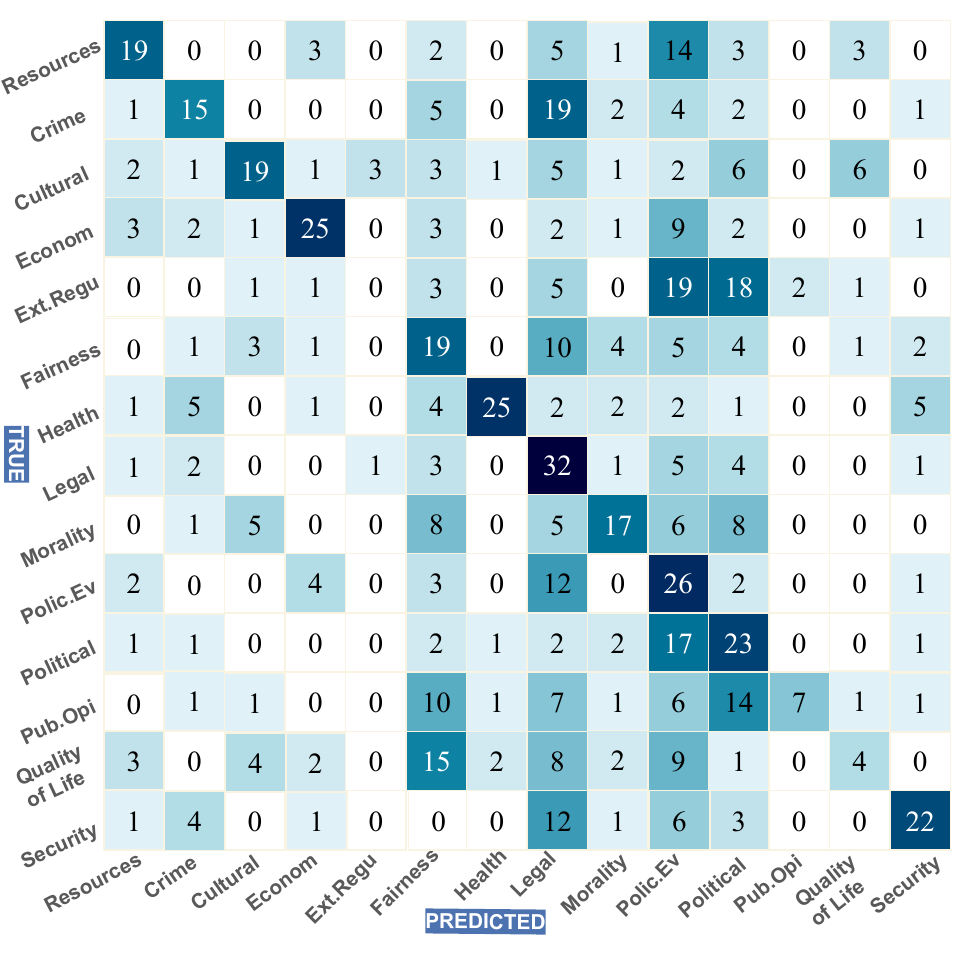}
  \caption{Confusion matrix of SCoT-based frame classification on the balanced dataset. Label names in this figure have been shortened for clarity; refer to Table \ref{tab:frame_definitions} for complete names.}
  \Description{The figure is a confusion matrix illustrating the performance of an SCoT-based frame classification model on a balanced dataset. It is presented as a square grid where the rows represent the ``TRUE'' frame categories and the columns represent the ``PREDICTED'' frame categories. The diagonal cells, running from the top-left to the bottom-right, are highlighted in a darker shade of blue, indicating correct classifications with higher numerical counts. Off-diagonal cells display the frequency of misclassifications between different frame pairs. The grid features shortened labels for various categories, such as ``Resources,'' ``Crime,'' ``Cultural,'' ``Econom,'' ``Ext.Regu,'' ``Fairness,'' ``Health,'' ``Legal,'' ``Morality,'' ``Polic.Ev,'' ``Political,'' ``Pub.Opi,'' ``Quality of Life,'' and ``Security.'' As noted in the caption, these labels are abbreviated for visual clarity, and readers are referred to Table 1 for the complete frame names}

  \label{fig:confusion-balanced}
\end{figure}

Figure~\ref{fig:confusion-balanced} visualizes the frame-level confusion matrix for the balanced dataset, aiding understanding of the model’s ``reasoning'' and failure patterns. While the diagonal is prominent, consistent misclassifications occur between conceptually adjacent categories like ``Fairness and equality'' and ``Legality'', or ``Policy prescription'' and ``Political''. These confusions reflect plausible interpretive ambiguity, not random errors.

For instance, 17 instances labeled “Policy prescription and evaluation” are predicted as “Political” (Figure~\ref{fig:confusion-balanced}). As per MFC definitions (Table~\ref{tab:frame_definitions}), the former focuses on policy solutions, while the latter on political processes. In immigration coverage, issues often carry both policy and political significance, creating ambiguity. We clearly acknowledge that some frame definitions are not mutually exclusive (e.g., policy prescription/political, legality), because they share certain common aspects.

Critically, these “errors” are more meaningful when considering the model’s SCoT output. Even when the label is incorrect, the justification often cites relevant keywords or logical ``reasoning'' that could persuade a human. Thus, misclassifications serve to evaluate the model’s ``reasoning'' for coherence, plausibility, and alignment with human interpretive practices, a key goal in explainable AI for human-centered tasks. 

\subsubsection{Structured ``Reasoning'': Interpretability and Extraction Efficiency}
To illustrate SCoT's performance, we use an example of the dataset whose headline can be paraphrased as: ``It’s not okay to let the people cross the border'', discussing a GPS device designed to assist undocumented immigrants crossing the U.S. border.

\textbf{Step-by-Step SCoT ``Reasoning'' (LLM Output):}
\begin{itemize}
    \item \textbf{Summarize the core message of the news text:}\\
    The news text discusses a GPS-enabled device created to help illegal immigrants cross the US-Mexico border safely, raising concerns about enabling unlawful activity. The author questions the motivations behind the device's creation and its potential impact on immigration policies.

    \item \textbf{Identify possible frames:}
    \begin{itemize}
        \item Frame 5: Legality, constitutionality, jurisprudence
        \item  Frame 8: Security and defense
    \end{itemize}

    \item \textbf{Compare and justify:}
    \begin{itemize}
        \item Frame 5: Legality, constitutionality, jurisprudence: This frame is the best fit because the news text revolves around the legality of the device and its potential impact on immigration policies. The author questions whether the device's creators are encouraging unlawful activity, and the device's encrypted nature to avoid electronic detection by the Border Patrol raises concerns about jurisprudence and the rule of law. Specific keywords like ``unlawful activity,'' ``illegal crossings,'' and ``civil disobedience'' support this frame.
        \item Frame 8: Security and defense: While this frame is applicable, it is less so than Frame 5. The news text does touch on security concerns, such as the device's encrypted nature and its potential to evade detection by the Border Patrol. However, the primary focus is on the legality and constitutionality of the device, making Frame 5 a more fitting choice. 
    \end{itemize}

    \item \textbf{Conclusion Frame:} \\
    Frame: 5. Legality, constitutionality, jurisprudence
\end{itemize}

This example demonstrates the SCoT approach: 
the model evaluates multiple possibilities, explains its ``reasoning'', and justifies its choice following a consistent structure, facilitating cross-article analysis. Key findings include:

\begin{itemize}
    \item From prediction to ``reasoning'': Unlike direct outputs from zero-shot or few-shot prompts, SCoT-generated responses reveal why a particular frame was chosen, thus providing traceable ``reasoning'' that users can verify, critique, or build upon.
    
    \item Data balance matters: Performance on imbalanced datasets may appear higher, but stems from dominant-frame bias. Balanced evaluation offers a more reliable view of ``reasoning'' quality and generalization.

    \item Making ambiguity visible: SCoT does not eliminate misclassifications, but it exposes the logic behind the process, surfacing competing interpretations and highlighting the subjective nature of many framing decisions.

    \item Suitability 
    for human-in-the-loop annotation: The structured format of SCoT allows users to inspect the model's logic, compare perspectives, and either confirm or challenge its interpretation. Rather than replacing human judgment, SCoT acts as an assistive tool, supporting collaborative annotation in domains where multiple readings are often valid.
\end{itemize}

These initial findings are then expanded 
in our next step: a human-centered evaluation to assess if SCoT-based ``reasoning'' is perceived as valid, invites reinterpretation, and supports, rather than replaces, human judgment with complementary perspectives.

\subsection{Human Evaluation of Structured ``Reasoning'' (RQ2)}

\subsubsection{Overall analysis of the annotations}

This subsection details the human annotation results, where 10 annotators evaluated SCoT-prompted LLM ``reasoning'' for frame classification. We focus on frame choice diversity (Q1), perceived ``reasoning'' validity (Q3), and ``reasoning'' persuasiveness (Q4). Q2 and Q5 are excluded here as they relate to annotator explanations and reconsideration, and will be referenced in the in-depth case analysis later in the section. 

\paragraph{Frame selection diversity (Q1).} 

Table \ref{tab:annotation_results} (a) shows unique frame choices per text ranging from 3 to 6, indicating diverse interpretations of the same news text. This suggests that longer texts (250 words) often contain multiple frames, leading to classification variability. While some texts (e.g., Text 4, Text 6) showed stronger alignment with the original dataset labels (7/10 annotators selecting the same frame), others (e.g., Text 7) had zero alignment. This variation underscores both the inherent subjectivity of frame classification in longer texts, and the fact that a single ``correct'' frame is not always apparent. This finding reinforces the view that agreement rate alone cannot capture the task's full complexity.

As we presented in the Data Section, the annotation process of the original data was carried out in three stages. In the first stage, the annotators had to annotate based on the definitions with minimal instructions, and independently. This is very similar to what we have done in our experiment, and the results are also similar. As \citeauthor{card2015media} mentioned \cite{card2015media}, in this first stage \textit{``The average number of frames identified in an article varied from 2.0 to 3.7 across annotators''} \shortcite{card2015media}, which means that at the beginning, the annotators said that there was more than one valid option, and that there was diversity in the selection. Subsequently, the authors implemented stage 2, where they assigned the same articles to pairs of annotators, in order to measure the subjectivity of the task. Then, in stage 3, there were also two annotators per article, and in cases of disagreement, they discussed until a consensus was reached. In our case, we did not deploy stages 2 and 3, as we focused on evaluating the ``reasoning'' ability of the LLM. What emerges as clear from these results is that the same text can be interpreted from multiple perspectives, even among individuals engaging with the same analytical task.

\begin{table*}[t]
\centering
\small
\begin{tabular}{cccccccccc}
\toprule
\multicolumn{3}{c}{\textbf{(a) Q1: Frame Selection}} & 
\multicolumn{3}{c}{\textbf{(b) Q3: ``Reasoning'' Validity}} & 
\multicolumn{4}{c}{\textbf{(c) Q4: ``Reasoning'' Persuasiveness}} \\
\cmidrule(r){1-3} \cmidrule(lr){4-6} \cmidrule(l){7-10}
\makecell{Text\\\#} & \makecell{Unique\\Frames\\(Q1)} & \makecell{Agree. with\\Dataset Frame\\(out of 10)} & 
\makecell{Text\\\#} & \makecell{Mean\\Q3} & \makecell{Std.\\Dev.} & 
\makecell{Text\\\#} & \makecell{Mean\\Q4} & \makecell{Std.\\Dev.} & \makecell{No Ans.\\Q4\\(out of 10)} \\
\midrule
1  & 4 & 3/10 & 1  & 4.3 & 1.16 & 1  & 3.67 & 1.51 & 4/10 \\ 
2  & 6 & 2/10 & 2  & 4.1 & 1.20 & 2  & 3.38 & 1.51 & 2/10 \\ 
3  & 4 & 1/10 & 3  & 4.1 & 0.99 & 3  & 2.40 & 0.55 & 5/10 \\ 
4  & 3 & 7/10 & 4  & 4.5 & 0.97 & 4  & 3.33 & 1.15 & 7/10 \\ 
5  & 4 & 2/10 & 5  & 3.6 & 1.26 & 5  & 3.00 & 1.49 & 0/10 \\ 
6  & 4 & 7/10 & 6  & 4.4 & 0.84 & 6  & 4.33 & 1.15 & 7/10 \\ 
7  & 3 & 0/10 & 7  & 3.1 & 1.29 & 7  & 2.10 & 1.20 & 0/10 \\ 
8  & 5 & 1/10 & 8  & 2.7 & 1.34 & 8  & 2.11 & 1.27 & 1/10 \\ 
9  & 5 & 3/10 & 9  & 4.3 & 0.95 & 9  & 4.00 & 1.00 & 3/10 \\ 
10 & 4 & 6/10 & 10 & 4.1 & 1.66 & 10 & 1.33 & 0.58 & 7/10 \\ 
\bottomrule
\end{tabular}
\caption{Annotation experimental data analysis: (a) Q1—Frame selection by annotators; (b) Q3—Structured reasoning validity; (c) Q4—Persuasiveness of the reasoning.}
\label{tab:annotation_results}
\end{table*}

\begin{figure}
  \centering
  \begin{subfigure}{0.45\textwidth}
    \centering
    \includegraphics[width=\linewidth]{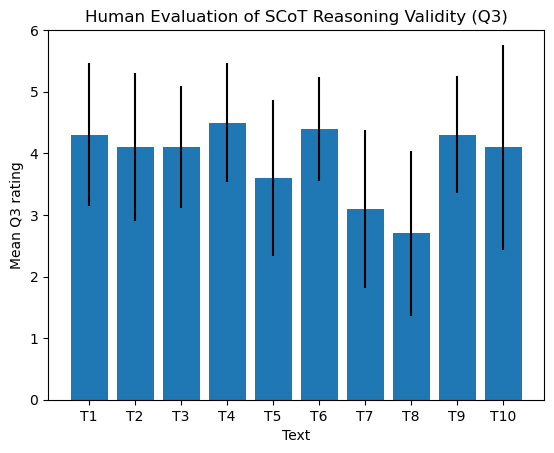}
    \caption{}
    \label{fig:e_length}
  \end{subfigure}
  \hfill
  \begin{subfigure}{0.45\textwidth}
    \centering
    \includegraphics[width=\linewidth]{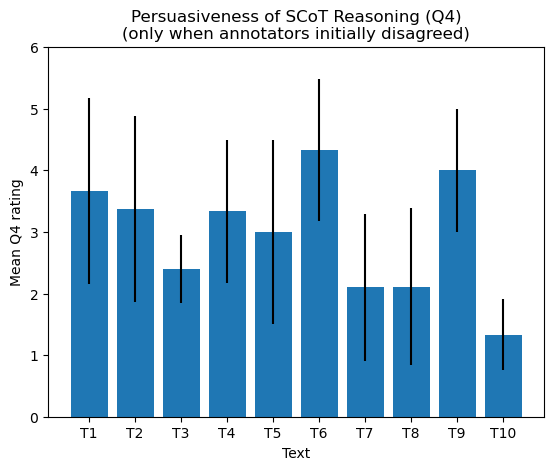}
    \caption{}
    \label{fig:n_length}
  \end{subfigure}
  \caption{Mean human ratings for (a) Q3: perceived validity of the SCoT reasoning, and (b) Q4: persuasiveness of the reasoning when annotators initially disagreed with the model’s frame.}
  \Description{Figure 4 displays two comparative bar charts, (a) and (b), evaluating human responses to SCoT (Structured Chain-of-Thought) reasoning across ten distinct text samples labeled T1 through T10. Chart (a) measures ``Perceived Validity (Q3),'' showing that the model’s reasoning generally maintains a consistent level of validity, with most samples scoring between 3 and 4.5; notable peaks are observed at T4 and T6, while T8 shows the lowest validity. Chart (b) evaluates ``Persuasiveness of Reasoning (Q4)'' in cases of initial annotator disagreement; this chart demonstrates higher variability, with sample T6 showing the strongest persuasive impact and sample T10 showing the least, suggesting that the model's ability to persuade human annotators is less uniform than its perceived validity. Each bar includes vertical error bars representing the variance in human ratings, highlighting that while some samples, like T8 in chart (a), have tighter agreement, others show significant divergence in human perception.}
  \label{fig:bars}
\end{figure}

\paragraph{``Reasoning'' validity (Q3).} 
Turning to part (b) of Table~\ref{tab:annotation_results}, the average Q3 score was 4.1 (out of 5), indicating that annotators generally found SCoT ``reasoning'' logical and coherent. High scores (e.g., Text 4: 4.5, Text 6: 4.4) suggest good reception when the model correctly identifies and justifies a dominant frame, whereas lower scores (e.g., Text 8: 2.7, Text 7: 3.1) imply misalignment with human expectations. Standard deviation (SD) values, also visualized in Figure~\ref{fig:bars} (a), reflect evaluation variability: lower SDs indicate strong consensus, while higher SDs suggest less clear or intuitive model reasoning. As an exploratory indicator, a non-parametric Friedman test suggests variation in perceived reasoning validity across texts ($\chi^2 = 19.52$, $p = 0.021$), though given the sample size of 10, this should be interpreted with caution and treated as a directional signal rather than a confirmatory finding.

\paragraph{Persuasiveness of ``reasoning'' (Q4).} 
Table \ref{tab:annotation_results} (c) shows the influence of the ``reasoning'' of SCoT on Q4 classification decisions. ``No Answer'' counts represent those annotators whose initial frame matched the LLM's output, indicating alignment with human intuition and reinforcing the model’s structured explanations. Mean Q4 scores, shown in Figure~\ref{fig:bars}(b), reveal how persuasive the reasoning was when the annotator initially disagreed with the model. High scores (e.g., Text 6: 4.33, Text 9: 4.0) suggest 
LLM influence, whereas low scores (e.g., Text 10: 1.33, Text 8: 2.11) indicate that annotators largely disregarded the explanation. Higher SDs (e.g., Text 2: 1.51) reflect varied responses, with some annotators finding the ``reasoning'' convincing and others not.


\subsubsection{In-Depth Case Analysis: Model–Human Agreement and ``Reasoning'' Impact.} To exemplify how SCoT ``reasoning'' interacts with human annotation in ambiguous contexts, we analyze the specific case of Text 3 (the GPS device article described previously.) Unlike prior examples that demonstrated 
acceptance or rejection of model ``reasoning'', Text 3 represents a mid-range case (Q3 mean: 3.7; Q4 mean: 2.4) where annotators diverged significantly in their interpretations. This case provides a nuanced view into how certain texts (longer than tweets or headlines, and with multiple embedded frames) can challenge both human and model classification.

\paragraph{Frame Diversity and Subjectivity (Q1–Q2).} Annotators showed diverse frame choices. One justified \textit{Morality} by focusing on the value of mobile devices for migrant emergencies, interpreting author feelings as an ethical debate on humanitarian aid. Another selected \textit{Legality}, referencing encrypted devices' implications for immigration authorities and focusing on law enforcement evasion. These examples highlight how annotators emphasize specific sentences leading to different frame conclusions, underscoring the subjective nature of frame selection.

\paragraph{Model ``Reasoning'' Validity (Q3) and Persuasiveness (Q4–Q5).}
The LLM classified the text under Legality (Frame 5), offering structured ``reasoning'' around legal language and phrases such as “unlawful activity” and “civil disobedience”. The Q3 average of 3.7 suggests that annotators generally found the model’s ``reasoning'' coherent, but not overwhelmingly persuasive.

In Q4, most annotators did not change their initial classification (low average: 2.4), but their Q5 justifications reveal important nuances. 
An annotator who did not change wrote: “I believe the ethical framing fits better because the main concern is about saving lives, not legal ramifications.”
In contrast, another annotator who did reconsider said:
“Although I first thought it was a moral issue, the SCoT explanation made me think more about the legal consequences of distributing such devices.”
These justifications underscore that SCoT ``reasoning'' surfaces alternative interpretations, even if it does not always persuade annotators to revise their choices.

\paragraph{Interpretation and Implications.}
Text 3 also illustrates broader patterns: annotators used overlapping quotes but interpreted them differently (legality, morality, public safety), confirming that frame definitions can be non-mutually exclusive. As \citeauthor{card2015media} acknowledged, this dataset involves inherent subjectivity and valid interpretative differences \cite{card2015media}, reinforcing the position that agreement metrics alone are insufficient for model evaluation. Even when SCoT ``reasoning'' does not change decisions, it often prompts reflection on overlooked aspects, thereby enhancing interpretability. The Text 3 example, reflecting ambiguous framing and mixed model reception, underscores the need for reasoning-aware evaluation in complex, high-stakes NLP tasks.\\

The annotation task revealed how SCoT ``reasoning'' from LLMs can indeed shape human interpretation in complex, multi-frame classification tasks. The variation in frame selections confirms that long-form policy texts often activate multiple interpretive lenses, and confirms 
that agreement metrics alone are insufficient: interpretability and transparency are essential, especially when human disagreement stems from textual ambiguity, not model failure. Additionally, the fact that annotators frequently assessed the SCoT ``reasoning'' as coherent and well-structured suggests that the value of SCoT lies not in enforcing agreement, but in providing a clear structure for comparing competing interpretations. As the creators of the dataset mentioned: \textit{``Disagreements continue to arise, however, reflecting the reality that the same article can cue different frames more strongly for different annotators. We view these disagreements not as a weakness, but as a source of useful information about the diversity of ways in which the same text can be interpreted by different audiences''} \cite{card2015media}, so the LLM could play an interactive role as an assistant to the annotator, one that can help annotators consider alternative perspectives in cases of disagreement, and reinforce their choice in case of agreement. As it was the case in stage 3 of the original data annotation process, where in case of disagreement, the two human annotators would argue until a consensus was reached, the LLM could be used 
as an “annotator” to argue with. We believe this to be a potentially relevant use because it could help the human annotator (a journalist or a person doing the frame analysis) to 
consider other perspectives, not as a replacement, but as an assistant, allowing the interaction between human and machine.

\section{Sociotechnical Implications and Responsible Use of LLMs in Media Analysis}
\label{sec: FATE}
The analysis of media framing is not only a methodological task but also a socially consequential practice, as the way events are framed can shape public understanding, political attitudes, and policy debates \cite{goffman1974frame}. This is particularly relevant in domains such as migration, where competing narratives can influence how vulnerable populations are perceived and discussed \cite{entman1993framing}. 

In this context, introducing LLMs into frame analysis  amplifies existing ethical concerns, with significant overlap with 
the risks identified in journalism and social sciences. For instance, both fields highlight issues of bias, transparency, and the potential for AI-generated content to mislead or manipulate audiences \cite{schroeder2025large, thapa2025large}. In journalism, specifically, the participatory co-design approach advocated by researchers emphasizes the importance of empowering journalists in the development and governance of LLMs to address challenges such as copyright infringement, declining public trust, and corporate control over AI tools \cite{suresh2024participation, tseng2025ownership}. This participatory model not only aligns with best practices for responsible AI, but also underscores the broader ethical imperative of aligning AI tools with the needs and values of the communities they serve.

In this section, we discuss how these concerns manifest in the specific context of frame analysis with LLMs, emphasizing 
the importance of responsible design practices \cite{tornberg2024best}.
A key concern is \textit{bias and fairness}. We acknowledge the inherent biases in LLMs, stemming from their training data, and consider how these biases might reflect prevalent viewpoints among news audiences \cite{alonso2024human}. While some scholars express concern about the black-box nature, subjectivity, unreliability, and potential bias of LLMs \cite{tornberg2024large}, it is important to recognize that human annotators also bring their own biases.  Our findings highlight that disagreements between human annotators and the LLM were not merely errors but often reflected alternative yet plausible interpretations. This suggests that LLM-assisted analysis does not simply reproduce bias, but can also surface competing perspectives, making the interpretative process more explicit while simultaneously requiring critical engagement from users.

In terms of \textit{transparency and accountability}, we emphasize the use of the open-source Llama3-8B model and of local deployment. This approach, as advocated by several researchers \cite{spirling2023open, rossi2024problems, alizadeh2023open}, increases the traceability of model outputs, provides greater control over the research process, and mitigates privacy concerns associated with sending sensitive data to external APIs \cite{tornberg2024large, ollion2024dangers}. This emphasis on transparency and local control ensures that researchers maintain control over their data and methods \cite{alizadeh2025open}. By transforming the LLM into an auditable assistant that surfaces alternative interpretations, our approach mitigates the risks of opaque automation and aligns with expert demands for human-in-the-loop validation and methodological openness \cite{wang2025media}.

Another relevant point is what Thapa et al. \shortcite{thapa2025large} called 
\textit{political and economic leaning}: ``LLMs might exhibit biases towards certain ideologies, potentially skewing research to favor particular viewpoints and influencing public opinion and policy decisions.'' Our findings, particularly the observation that annotators sometimes revised their judgments based on the LLM ``reasoning'', underscore this risk. While this behavior suggests a potential assistive role for LLMs, it also highlights the vulnerability of human judgment to LLM-generated framing, raising concerns about the potential misuse of these systems in shaping public opinion. In contexts such as migration, where framing can influence societal attitudes toward marginalized groups, these systems may be used not only for analysis but also for narrative shaping. This risk is far from theoretical; recent work by Wang et al. \shortcite{wang2025detecting} and Khatiwada et al. \shortcite{khatiwada2026ai} has demonstrated how effectively LLMs can alter news perceptions through framing shifts, underscoring the need for deeper study of human-AI collaborative settings.

Finally, we address the consideration of \textit{access to computational resources}. Recognizing that LLM applications are resource-intensive, we employed a model that can be run locally on a single GPU. By using a locally deployable model, we address the ethical concern of equitable access to advanced LLM tools, ensuring that researchers and journalists with limited resources are not excluded from benefiting from these technologies.

\section{Limitations and Future Work}
\label{sec: Limitations}

Our study has several limitations, which also represent directions for future work.

1. Topic and dataset size. Our experiments focused on news about immigration as a case study, using a subset of 700 articles that allowed for a controlled experimental protocol. We consider this dataset large enough to test our frame analysis approach. Since the structured approach of summarizing, identifying frames, and providing evidence-based justifications is topic-independent, future work could extend the analysis to other socially relevant topics within the MFC dataset, such as climate change, to demonstrate the generalizability of our proposed method.

2. Small annotation sample.
Our annotation sample was small compared to \citeauthor{card2015media}'s extensive process \shortcite{card2015media}. While our primary aim was to evaluate LLM argument coherence and interpretability rather than solely frame annotation, future work could increase the number of evaluated texts and SCoT outputs.

3. Overlap and ambiguity between frame categories.
Several frame definitions in our taxonomy are not mutually exclusive. Some frames, like “Morality,” are also more abstract and harder to identify, compared to 
other frames like “Economic.” These ambiguities make it challenging for both humans and models to assign a single correct label, especially for texts that could plausibly fit multiple frames; in contrast, the MFC dataset was annotated with one single ``primary frame''. In the future, it would be interesting to create a multi-label dataset and test the SCoT approach we proposed.

These limitations point to concrete directions for future work: including using datasets from other topics, involving more annotators, and considering multi-label or consensus-based approaches for frame analysis.

\section{Conclusion}
\label{sec: conclusion}
This study investigated how SCoT prompts can make frame classification more transparent and practical, using open-source LLMs. To conclude, we answer the two Research Questions that we posed: 

RQ1: To what extent can structured prompting strategies provide interpretable frame classification with resource-constrained LLMs, and how does SCoT facilitate this process?
By using a step-by-step prompting strategy 
that runs locally on a single GPU, the performance varies across balanced and imbalanced datasets, highlighting the need to interpret “agreement rate” cautiously in subjective tasks, and relying more on metrics adapted to imbalanced data such as F1. It should be noted that the output of the SCoT prompt facilitates human interpretation and reinforces the role of the LLM as an interactive assistant, where the person doing frame analysis can understand the LLM ``reasoning'', in addition to improving results compared to zero-shot and few-shot settings.

RQ2: How do human evaluators perceive and respond to the model ``reasoning'' in SCoT-based frame classification, particularly in terms of agreement and confidence, given the task's inherent subjectivity? Through a human evaluation study, we found that most annotators found the model’s ``reasoning'' logical (average Q3 score: 4.1), and occasionally persuasive enough to change their initial frame selection.  Even when they disagreed, annotators often still perceived the ``reasoning'' as meaningful. This suggests that SCoT outputs can help surface new perspectives, clarify ``reasoning'' paths, and potentially support more thoughtful human annotation.

By prioritizing local deployment, interpretability, and human oversight, this work offers a concrete contribution to the broader goal of developing AI tools that are accessible and accountable in socially sensitive research domains, where responsible computational analysis of media narratives intersects with broader societal objectives around migration and public access to information defined by the United Nations.

\begin{acks}
We thank all participants in the evaluation process. This work was supported by the EU Horizon Europe program through the ELIAS
project (No. 101120237)
\end{acks}

\bibliographystyle{ACM-Reference-Format}
\bibliography{sample-base}


\end{document}